\documentclass{article}

\usepackage[preprint]{neurips_2022}
\usepackage{graphicx}
\usepackage{graphics}
\usepackage{subcaption}
\usepackage{multirow}

\usepackage[utf8]{inputenc} 
\usepackage[T1]{fontenc}    
\usepackage{hyperref}       
\usepackage{url}            
\usepackage{booktabs}       
\usepackage{amsfonts}       
\usepackage{nicefrac}       
\usepackage{microtype}      
\usepackage{xcolor}         
\usepackage{listings}
\usepackage{color}

\definecolor{dkgreen}{rgb}{0,0.6,0}
\definecolor{gray}{rgb}{0.5,0.5,0.5}
\definecolor{mauve}{rgb}{0.58,0,0.82}

\usepackage[ruled,vlined]{algorithm2e}
\usepackage{algpseudocode}
\SetKwInput{kwInput}{Input}
\SetKwInput{kwOutput}{Output}
\SetKw{Break}{break}
\usepackage{float}
\usepackage{amsmath}

\DeclareMathOperator*{\argmin}{arg\,min}

\usepackage{blindtext}
\usepackage[most]{tcolorbox} 
\definecolor{block-gray}{gray}{0.95}

\newtcolorbox{zitat}[2][]{%
    colback=block-gray,
    grow to right by=-10mm,
    grow to left by=-10mm, 
    boxrule=0pt,
    boxsep=0pt,
    breakable,
    enhanced jigsaw,
    borderline west={4pt}{0pt}{gray},
    title={#2\par},
    colbacktitle={block-gray},
    coltitle={black},
    fonttitle={\large\bfseries},
    attach title to upper={},
    #1,
}

\title{
    
    HyperTuning: 
    Toward Adapting Large Language Models without Back-propagation
}

\author{%
  Jason Phang\textsuperscript{1,2}\thanks{Work done at Microsoft. Correspondence: jasonphang@nyu.edu}\ ,
  Yi Mao\textsuperscript{3},
  Pengcheng He\textsuperscript{3},
  Weizhu Chen\textsuperscript{3}\\
  \ \\
  \textsuperscript{1}New York University
  \\
  \textsuperscript{2}EleutherAI
  \\
  \textsuperscript{3}Microsoft Azure AI
}

\begin{document}

\maketitle

\begin{abstract}


Fine-tuning large language models for different tasks can be costly and inefficient, and even methods that reduce the number of tuned parameters still require full gradient-based optimization.
We propose HyperTuning, a novel approach to model adaptation that uses a hypermodel to generate task-specific parameters for a fixed downstream model.
We demonstrate a simple setup for hypertuning with HyperT5, a T5-based hypermodel that produces soft prefixes or LoRA parameters for a frozen T5 model from few-shot examples.
We train HyperT5 in two stages: first, hyperpretraining with a modified conditional language modeling objective that trains a hypermodel to generate parameters; second, multi-task fine-tuning (MTF) on a large number of diverse language tasks.
We evaluate HyperT5 on P3, MetaICL and Super-NaturalInstructions datasets, and show that it can effectively generate parameters for unseen tasks.
Moreover, we show that using hypermodel-generated parameters as initializations for further parameter-efficient fine-tuning improves performance.
HyperTuning can thus be a flexible and efficient way to leverage large language models for diverse downstream applications.

\end{abstract}
\section{Introduction}


While language models (LMs) have achieved remarkable capabilities with increasing model size \citep{brown2020gpt3,chowdery2022palm},
fine-tuning them on specific downstream tasks introduces significant engineering challenges and computational costs.
Although large models can perform zero-shot, instruction-prompted, and few-shot learning \citep{sanh2022t0,wei2022flan}, they are usually outperformed by fully fine-tuned models when sufficient training data is available.


To reduce the computational and memory overhead of fine-tuning LMs, parameter-efficient fine-tuning (PEFT) methods have been proposed, such as adapters \citep{houlsby2019adapters}, prefix tuning \citep{li-liang-2021-prefix}, and prompt tuning \citep{lester2021prompt}.
These methods update only a small subset of (possibly new) parameters of the LM, and have achieved competitive performance with full fine-tuning \citep{ding2022delta}.
However, PEFT methods still require full back-propagation through the LM during training, which is computationally expensive and memory intensive.
Given that (1) only a small number of parameters need to be updated to adapt an LM to a given task, (2) very large LMs have demonstrated strong in-context learning capabilities on a forward pass, and (3) a forward pass for very large LMs already entails a substantial amount of computation, we hypothesize that it is possible to train a separate model to perform the optimization or adaptation procedure entirely, using only a forward pass.


To avoid the costly computation of back-propagating through the LM to produce the parameter updates, especially for thousands or millions of iterations during training, we propose a new paradigm of \textbf{hypertuning}: using a \textit{hypermodel} to adapt a \textit{downstream} LM to a desired application.
As a concrete proof of concept, we explore a simple setup where hypermodels take as input a set of few-shot examples from a given task, and output the PEFT parameters corresponding to that task in a single forward pass.

To demonstrate the feasibility of this approach, we train \textit{HyperT5}: a set of T5-based hypermodels that output soft prefixes \citep{li-liang-2021-prefix} or LoRA parameters \citep{hu2022lora}, to be incorporated into a frozen downstream T5 LM.
To train HyperT5, we introduce a two-stage procedure for training hypermodels: \textit{hyperpretraining}, where we adapt a pretrained LM to generate PEFT parameters via a modified language modeling objective, followed by \textit{multi-task fine-tuning} (MTF) the hypermodel.
After training, HyperT5 models can take few-shot examples from unseen tasks and generate the corresponding PEFT parameters, allowing us to adapt a downstream LM without back-propagation.
We show in experiments across P3, Super-NaturalInstructions and MetaICL datasets that LMs can be hypertuned using just a small number of examples.
Furthermore, we show that when the hypermodel-generated parameters are used as initializations for further parameter-efficient fine-tuning, we can achieve faster training convergence and better overall performance.


This work serves as a first step toward hypertuning, and we are are aware of certain limitations of this preliminary setup.
Because our current formulation of hypermodels can only take a small number of examples as input, its performance cannot compare to full parameter-efficient fine-tuning or full fine-tuning.
HyperT5 also generally underperforms T5 explicitly trained for few-shot in-context learning with full attention across examples, although we note that the latter is more computationally expensive to use at inference time.
Nevertheless, we believe that our results demonstrate a promising step toward model adaptation without the need for back-propagation.

We plan to release the code and model weights for \textbf{HyperT5}, as well as the multi-task fine-tuned versions for the three datasets listed above.


\section{Related Work}

\paragraph{HyperNetworks}
Several works have explored the concept of "hypernetworks," where an auxiliary network is used to generate parameters for a primary network. This terminology was first introduced by \citet{ha2017hypernetworks} and applied to LSTMs. Among Transformer-based language models, \citet{mahabadi2021hyperformer} and \citet{he2022hyperprompt} incorporated hypernetworks into T5 models for knowledge sharing during multitask fine-tuning. \citet{peebles2022gdotpt} utilized a Transformer with diffusion for generating full model parameters for image-recognition and Cartpole tasks. Similarly, \citet{lester2022recycling} trained models to generate soft prompts for transferring between downstream models.
Our work is closely related to \citet{budhaditya2022boosting}, who also used a hypernetwork to modify downstream model parameters and incorporated Super-NaturalInstuctions (S-NI) in their experimental setting. They found that incorporating instructions via a hypernetwork trained with MAML \citep{finn2017maml} improved downstream performance.

\paragraph{Multi-task Training and Transfer}
A crucial ingredient to hypertuning is the transferrability of task knowledge and generalization to novel tasks.
Many past works \citep{phang2018stilts,pruksachatkun-etal-2020-intermediate,vu2020exploring} have explored the effectiveness of single- and multi-task transfer learning.
More recent work has shown that large-scale multi-task training tends allows models to generalize to unseen tasks \citep{sanh2022t0,wei2022flan,wang2022sni,chung2022flant5}.
\citet{min-etal-2022-metaicl} and \citet{chen-etal-2022-meta} show that few-shot learning also benefits from multi-task training. \citet{pfeiffer2020adapterfusion}, \citet{vu2021spot} and \citet{gu2021ppt} have also explored transfer learning among PEFT methods.


\section{HyperTuning}
\label{sec:hypertuning}

\begin{figure}
  \centering
  \includegraphics[width=\linewidth]{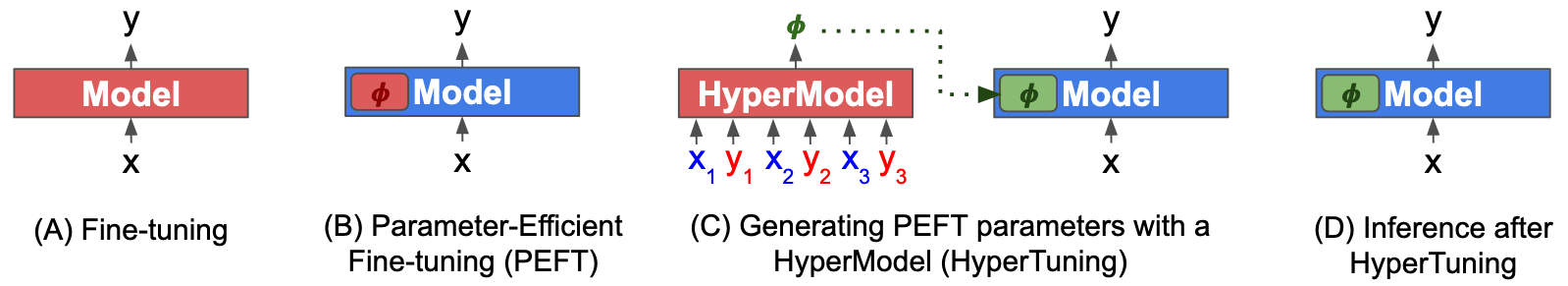}
  \caption{
  Overview of HyperTuning.
  (A) Fine-tuning, where all model parameters are updated (red).
  (B) Parameter-efficient fine-tuning (PEFT), where all model parameters are frozen (blue) and only a small number of parameters, $\phi$, are updated.
  (C) HyperTuning, where a hypermodel is used to generate parameters $\phi$ for a frozen downstream model.
  For instance, a hypermodel may take a set of few-shot examples to determine what $\phi$ to generate.
  Only the hypermodel's parameters are updated during training.
  (D) At inference time, the parameters $\phi$ only need to be generated once, and thereafter only need to store $\phi$, with no need to retain the few-shot examples.
  }
\label{fig:summary_plot}
\end{figure}

The impetus for using hypermodels for adapting downstream models derives from two recent developments in natural language processing: 


\paragraph{1) Large language models can perform in-context learning effectively.}
Large language models have been shown to be able to learn from the context of a small number of examples or instructions for a task, without any prior training on that task \citep{brown2020gpt3,min-etal-2022-metaicl,wang2022sni}.
This suggests that models can ``understand'' what the task is and how to tackle it based on a few samples or descriptions of the task.
This capability appears to improve as the models get larger or are trained on more relevant data \citep{chowdery2022palm,ouyang2022instructgpt,bai2022helpful}.


\paragraph{2) Large language models can be adapted to downstream tasks by tuning a small set of parameters.}
Along with the growth in model sizes, there have been significant advances in fine-tuning methods that only modify a small number of parameters (possibly adding some new ones) in a frozen language model to adapt it to a specific task \citep{houlsby2019adapters,li-liang-2021-prefix,lester2021prompt,ding2022delta}.
These methods often achieve performance comparable to fine-tuning all parameters in the model.
Importantly, the number of parameters that need to be changed is small enough that it is feasible to train a model to generate them \citep{qin2021ipt,lester2022recycling}.


Taken together, these findings suggest that we may be able to use an auxiliary model that can first extract some task-relevant knowledge from some input that describes the task (e.g. instruction, few-shot examples), and then generate a small number of adaptive parameters, thereby changing the main model's behavior to suit the task.
This approach, if successful, would enable us to adapt models to downstream applications without using backpropagation, or storing the encoded representations of few-shot examples in memory.
In other words, we can delegate the work of model adaptation to a separate model.



We call this approach \textbf{hypertuning}, inspired by the work on hypernetworks by \citet{ha2017hypernetworks}.
Hypertuning uses a \textit{hypermodel} to adapt a \textit{downstream model} to a target downstream task or application.
This is differs from \textit{fine-tuning}, which uses backpropagation and a gradient descent algorithm to update model parameters.
In this work, we present one possible formulation of hypertuning using few-shot examples and generating a small set of parameters with a single forward pass through the hypermodel.
However, this is just one possible way of performing hypertuning, and the idea of adapting models with hypermodels can be generalized to many other cases.
For example, hypermodels could also be trained to predict gradients or generate parameter updates based on input-output pairs.
This way, hypermodels could work with large training sets, not just a few examples.
Ultimately, with sufficiently general and well-trained hypermodels, we may be able to replace gradient-descent-based fine-tuning pipelines with hypertuning for many applications, while achieving similar or better performance.


\subsection{HyperTuning with Fewshot Examples}


Let $M$ be a model with parameters $\theta$, initialized at $\theta_0$ from pretraining, and $\mathbb{L}$ a loss function.
Given a dataset of size $N$ with input-output pairs $\{(x,y)\}$, standard fine-tuning minimizes the following objective over $\theta$:

\begin{equation}
    \argmin_{\theta}{\frac{1}{N}\sum_{\{(x,y)\}}{\mathbb{L}\Big(y, M(\theta;x)}\Big)}
\end{equation}


In the case of parameter-efficient fine-tuning (PEFT), we fix $\theta=\theta_0$ and introduce a small set of trainable parameters $\phi$ (e.g. adapter parameters, soft prompts) that are injected into $M$.
We optimize only over $\phi$:

\begin{equation}
    \argmin_{\phi}{\frac{1}{N}\sum_{\{(x,y)\}}{\mathbb{L}\Big(y, M(\theta_0;x,\phi)}\Big)}
\end{equation}


For hypertuning, we further define a \textit{hypermodel} $H$ with parameters $\xi$ that produces PEFT parameters $\hat{\phi}$ based on its input, which can be a set of few-shot examples or task instructions.
For example, if the hypermodel input is a set of few-shot examples $\{(x_i, y_i)\}_K$, we have:

\begin{equation}
    \label{eq:hypermodel}
    \hat{\phi} = H\Big(\xi; \{(x_i, y_i)\}_K\Big)
\end{equation}


One way to train the hypermodel $(H, \xi)$ is to perform PEFT on many tasks and use the resulting $\phi$ as targets.
However, this is costly in computation, requiring many fine-tuning runs, and does not leverage cross-task knowledge transfer.
Instead, we propose to train the hypermodel end-to-end, optimizing through the frozen model $(M, \theta_0)$.
Hence, the hypermodel training objective is:

\begin{equation}
    \argmin_{\xi}{\frac{1}{N}\sum_{\{(x,y)\},\{\{(x_i, y_i)\}_K\}}{\mathbb{L}\bigg(y, M\Big(\theta_0;x,H(\xi; \{(x_i, y_i)\}_K)\Big)}\bigg)}
\end{equation}


At each training step, we sample a \textit{target example} $(x,y)$ and non-overlapping few-shot examples $\{(x_i, y_i)\}_K$.
We generate $\hat{\phi}$ from the few-shot examples and compute the loss with respect to $(x,y)$ and $\hat{\phi}$.
We then back-propagate the gradients through both $M$ and $H$ to update $\xi$.


Note that since $\hat{\phi}$ does not depend on $x$, it can be computed once for a given set of few-shot examples and reused for downstream predictions.
At inference time, we can use $\hat{\phi}$ directly without storing or recomputing the representations for $\{(x,y)\},\{(x_i, y_i)\}_K$, saving memory and computation.\footnote{By construction, few-shot examples occupy at least K times the memory of the target input $x$.}

\section{HyperT5: A T5-Based HyperModel}

\subsection{Architecture and Setup}

\begin{figure}
  \centering
  \includegraphics[width=\linewidth]{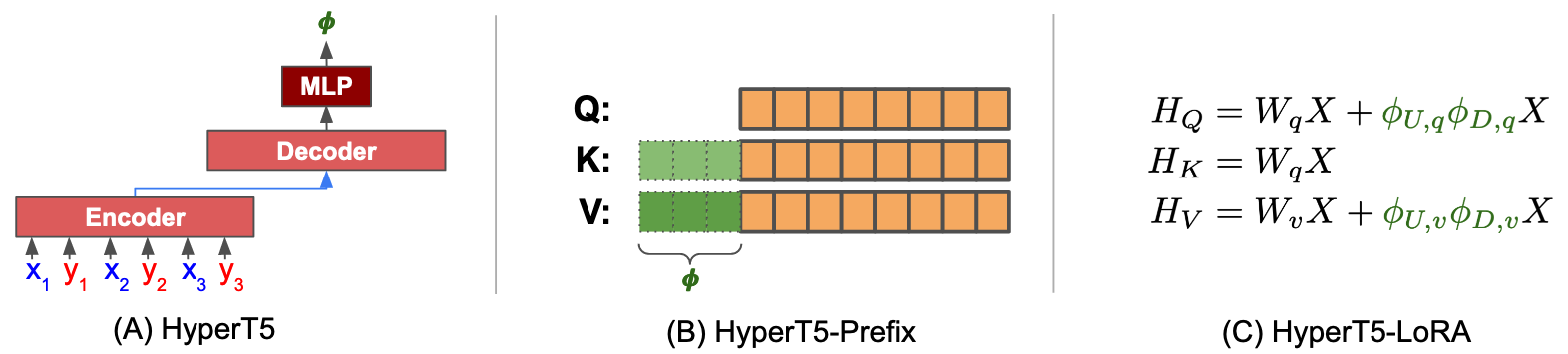}
  \caption{
    Overview of HyperT5.
    (A) HyperT5 takes as input few-shot examples and outputs PEFT parameters $\phi$.
    The model is initialized from an LM-adapted T5.
    (B) In HyperT5-Prefix, $\phi$ are key and value prefixes for every attention layer.
    (C) In HyperT5-LoRA, $\phi$ are additive low-rank modifications to the query and value linear maps.
  }
\label{fig:hypert5}
\end{figure}

To demonstrate the feasibility of hypertuning, we propose \textit{HyperT5}, a hypermodel based on T5, where both the hypermodel and the downstream model share a T5 backbone (Figure~\ref{fig:hypert5}A).
We use a frozen LM-adapted T5 \footnote{This is the model introduced by \citet{lester2021prompt}. We use the T5 v1.1 architecture and initialize all experiments with the LM-adapted parameters, unless stated otherwise.} as the downstream model.
The hypermodel is also initialized with LM-adapted T5 parameters, but with some architectural changes.
As defined in Equation~\ref{eq:hypermodel}, the hypermodel encoder takes  the few-shot examples (and/or task definitions, in the case of S-NI) as input.
The hypermodel decoder takes a fixed set of newly learned token embeddings as input, and outputs a set of decoder token representations, which are then fed to a set of MLPs to generate the PEFT parameters $\phi$ for the downstream model.
We also remove the causal masking from the decoder, since the hypermodel does not perform autoregressive generation. 

We experiment with two PEFT methods: prefix tuning \citep{li-liang-2021-prefix} and LoRA \citep{hu2022lora}.
Prefix tuning (Figure~\ref{fig:hypert5}B) prepends a set of learned key and value representations within each attention layer, while LoRA (Figure~\ref{fig:hypert5}C) learns a low-rank additive modification to the query and value linear maps.
Both PEFT methods have been shown to achieve good performance across a wide range of tasks \citep{ding2022delta}.
\citet{chan2022differently} also suggest that modifying in-context representations and model weights can lead to different model behaviors, and we seek to demonstrate that hypertuning is applicable to very different PEFT methods.
We name the respective hypermodels \textit{HyperT5-Prefix} and \textit{HyperT5-LoRA}.

The number of decoder input tokens and the size of the MLPs  depend on the choice of PEFT method and its hyperparameters.
For example, for HyperT5-Prefix that generates soft prefixes corresponding to prefix tuning,  $\phi$ will be of the shape $[L,2,2,P,H]$, 
where $L$ is the number of layers, 2 is for the encoder and decoder, 2 is for the key and value prefixes, $P$ is the number of prefix tokens, and $H$ is the hidden size. 
We set the number of decoder input tokens to be $2P$.
We provide pseudo-code for HyperT5-Prefix and HyperT5-LoRA models in the Figure~\ref{app:pseudoprefix} and Figure~\ref{app:pseudolora} in the Appendix.


\subsection{HyperPretraining}

To train HyperT5, we first undergo an additional stage of pretraining to adapt the hypermodel to generate parameters $\phi$ for the downstream model, which we call \textit{hyperpretraining}.
As we show in Section~\ref{sec:hyperpretrainingresults}, hyperpretraining is crucial for good hypermodel performance.

\begin{figure}
  \centering
  \includegraphics[width=\linewidth]{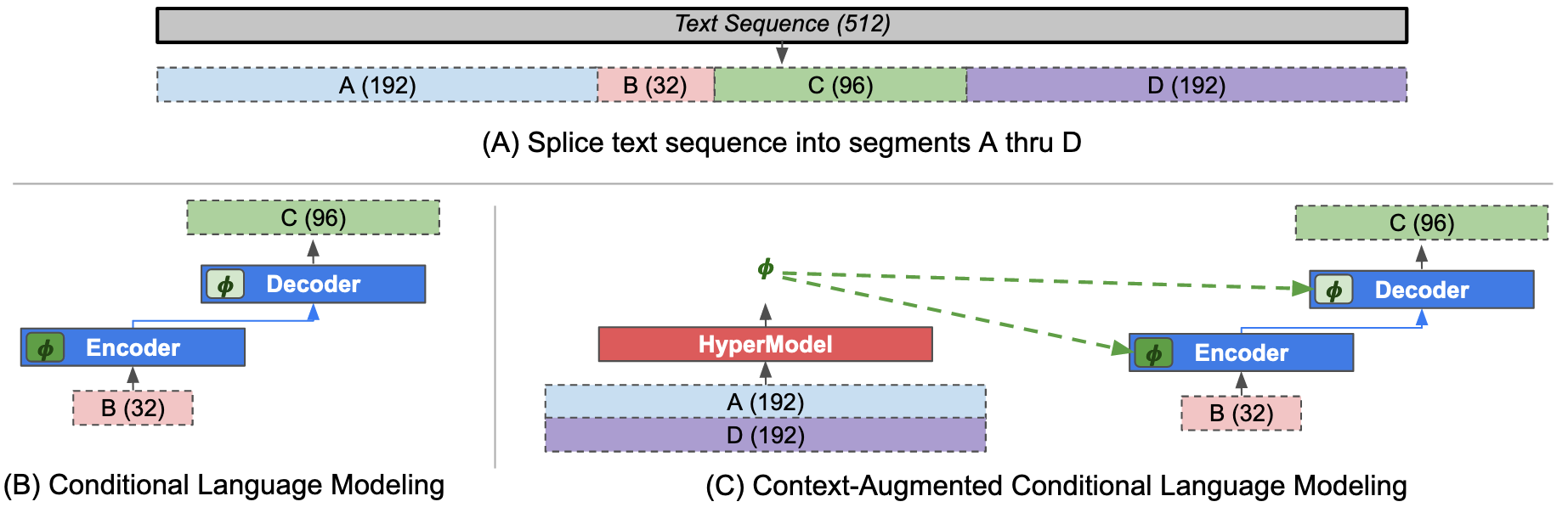}
  \caption{
    Overview of HyperPretraining using the Context-Augmented Conditional Language Modeling (CACLM) objective to train a hypermodel to predict PEFT parameters $\phi$.
    (A) Sample a sequence of 512 tokens from a pretraining corpus, and splice into 4 segments A--D.
    (B) The frozen downstream model takes as input B and predicts continuation C.
    (C) The hypermodel is trained to encode additional context A and D into PEFT parameters $\phi$, providing additional information to the downstream model to predict C.
  }
\label{fig:hyperpretraining}
\end{figure}

We propose a simple scheme for hyperpretraining using a \textit{Context-Augmented Conditional Language Modeling} (CACLM) objective, which extends the conditional language-modeling (CLM) objective of T5 LM-adaptation.
As shown in Figure~\ref{fig:hyperpretraining}, we sample a 512-token sequence from a pretraining corpus and split it into four consecutive segments A--D.
The downstream model receives segment B as input and predicts segment C, following the CLM objective.
The hypermodel receives segments A and D as input, which provide additional context from the same document, and outputs PEFT parameters for the downstream model.\footnote{Segments A and D are marked by sentinel tokens.} 
The hypermodel thus compresses contextual information to assist the downstream model in its CLM task.
We also make segment B very short (32 tokens) to encourage the downstream model to depend on the hypermodel information for accurate prediction of tokens in C.

During hyperpretraining, we freeze the downstream model and only update the hypermodel parameters, training for 100K steps on the C4 dataset \citep{raffel2020t5}.
We perform hyperpretraining separately for HyperT5-Prefix and HyperT5-LoRA models.
Hyperparameters can be found in Appendix~\ref{app:trainingdetails}.

\section{Multi-Task Fine-Tuning with HyperT5}

\subsection{Multitask Fine-Tuning (MTF)}


After hyperpretraining, we conduct a second stage of training to train the hypermodel to generate task-specific PEFT parameters based on a small number of examples that we provide as input (Figure~\ref{fig:summary_plot}C).
By performing multi-task fine-tuning on a sufficiently large number of tasks, we hope to have the hypermodel learn to generalize to generate parameters for unseen tasks.
We adopt a similar training setup to MetaICL \citep{min-etal-2022-metaicl}, which uses multi-task fine-tuning \citep{sanh2022t0, wei2022flan} with both a target input example ($x$) and a set of few-shot input-output pairs $\{(x_i,y_i)\}_K$ as inputs.
The hypermodel takes the few-shot pairs as input, while the downstream model takes the target example as input, as shown in Equation~\ref{eq:hypermodel}.
We fine-tune only the hypermodel parameters and keep the downstream model parameters fixed, unless otherwise stated.
Appendix~\ref{app:inputformatting} shows how we format the few-shot inputs.


We compare our approach with two baselines: multi-task fine-tuning of a T5 model without few-shot inputs, and MetaICL (multi-task fine-tuning with few-shot inputs).
In MetaICL, the few-shot pairs are concatenated with the target example as input, both during training and evaluation on new tasks. 
We also include baselines that use PEFT methods for multi-task fine-tuning, i.e. learning a single set of prefix tuning or LoRA parameters.


We perform multi-task fine-tuning for 10,000 steps with a batch size of 256.
For models that use few-shot inputs (MTF with fewshot, and hypermodels), we use up to 16 examples, and truncate tokens that exceed the maximum input length.
Appendix~\ref{app:datadetails} provides more details on the datasets.

\subsection{Datasets}

To demonstrate the generality of our approach, we conduct experiments on three different multi-task training datasets, each with different held-out tasks and evaluation protocols.

\textbf{Public Pool of Prompts (P3)} \citep{sanh2022t0} consists of 62 task datasets, and was used in training the T0 models. 
The prompt are formatted with 0-shot inference in mind, and often contain instructions or the possible answer options.
For training our models, we use the T0-train subset. In order to fit multiple examples into the hypermodel's context, we further exclude dataset-prompt subsets with average input sequence lengths longer than 320 tokens. The list of included dataset-prompts can be found in Figure~\ref{fig:appp3train}.
Evaluation is performed on a fixed set of held-out tasks, based on multiple-choice scoring with accuracy.
We exclude StoryCloze from evaluation as the task is not distributed with training data.

\textbf{MetaICL} \citep{min-etal-2022-metaicl} introduced a few-shot multi-task training dataset, which is an extension of CrossFit \citep{ye-etal-2021-crossfit} with UnifiedQA \citep{khashabi-etal-2020-unifiedqa} and the addition of training data.
For brevity, we will refer to this dataset as MetaICL.
Unlike P3 and S-NI, the task inputs are not formatted for 0-shot inference; for instance, the task inputs may give no clue as to the goal of the task, or what the output space is.
They provide several different train-task splits for tasks, of which we run our experiments on three (HR$\rightarrow$LR, Non-NLI$\rightarrow$NLI, Non-Class$\rightarrow$Class) to economize on computation costs.
Evaluation is performed on held-out tasks, with ROUGE or Macro-F1 on model generations depending on the task.

\textbf{Super-NaturalInstructions (S-NI)} \citep{wang2022sni} consists of over 1,600 task datasets, each with a task definition as well as a fixed set of positive and negative demonstrations.
Following their findings, we focus our experiments on two settings: using only the task definition as the hypermodel input, and using definitions alongside two fixed positive examples.
We only use the English tasks within the dataset.
Evaluation is performed on a set of held-out tasks using ROUGE-L on model generations.

\subsection{Results}

\label{sec:results}

\subsubsection{P3}

\label{sec:results_p3}

\begin{table}[th]
\centering
\small
\begin{tabular}{lrrrrrrrrrrr}
    \toprule
    & ANLI & HSwag & CB & COPA & RTE & WiC & WSC & WGD & \underline{AVG}
    \\ \midrule
        \multicolumn{10}{l}{\textit{Full Fine-Tuning}} \\
    \ \ T5-MTF
    & 33.4 & 28.0 & 63.0 & 77.9 & 71.1 & 50.8 & 61.0 & 53.4 & 54.8
    \\
    \ \ T5-MTF-Few-shot
    & 35.3 & 27.5 & 68.6 & 70.5 & 75.2 & 51.7 & 62.1 & 52.2 & 55.4
    \\ \midrule
      \multicolumn{10}{l}{\textit{Parameter-Efficient Fine-Tuning (PEFT)}} \\
    \ \ T5-MTF (Prefix)
    & 33.1 & 26.1 & 53.9 & 67.8 & 60.5 & 49.8 & 54.7 & 51.4 & 49.7
    \\
    \ \ T5-MTF (LoRA)
    & 32.9 & 26.0 & 36.0 & 59.7 & 49.8 & 51.2 & 58.1 & 50.5 & 45.5
    \\ \midrule
      \multicolumn{10}{l}{\textit{HyperTuning}} \\
    \ \ HyperT5-Prefix
    & 33.4 & 32.3 & 60.1 & 73.9 & 71.5 & 51.1 & 63.0 & 51.1 & 54.6
    \\
    \ \ HyperT5-LoRA
    & 33.6 & 33.0 & 49.5 & 74.2 & 67.4 & 52.0 & 64.0 & 52.9 & 53.3
    \\ \midrule
      \multicolumn{10}{l}{\textit{HyperTuning + Fine-Tuning}} \\
    \ \ HyperT5-Prefix+
    & 34.5 & 32.2 & 58.1 & 78.4 & 76.5 & 50.4 & 63.8 & 54.3 & 56.0
    \\
    \ \ HyperT5-LoRA+
    & 33.9 & 30.7 & 62.1 & 75.8 & 72.3 & 50.8 & 64.6 & 54.5 & 55.6
    \\ \bottomrule
\end{tabular}%
\caption{
  Results on P3 on held-out tasks (dev) with T5-Large models.
  T0 results taken from \citet{sanh2022t0}.
}
\label{tab:table_01_p3}
\end{table}
\begin{table}[th]
\centering
\small
\begin{tabular}{lrrrrrrrrrrr}
    \toprule
    & ANLI & HSwag & CB & COPA & RTE & WiC & WSC & WGD & \underline{AVG}
    \\ \midrule
        \multicolumn{10}{l}{\textit{Full Fine-Tuning}} \\
    \ \ T5-MTF
    & 39.9 & 29.4 & 64.5 & 88.0 & 80.8 & 51.7 & 60.7 & 57.9 & 59.1
    \\
    \ \ T5-MTF-Few-shot
    & 37.9 & 30.9 & 67.6 & 90.5 & 76.6 & 51.2 & 63.3 & 61.1 & 59.9
    \\ \midrule
        \multicolumn{10}{l}{\textit{Parameter-Efficient Fine-Tuning (PEFT)}} \\
    \ \ T5-MTF (Prefix)
    & 38.3 & 31.2 & 61.4 & 82.4 & 78.6 & 52.6 & 57.0 & 54.3 & 57.0
    \\
    \ \ T5-MTF (LoRA)
    & 33.9 & 26.4 & 47.1 & 67.2 & 53.3 & 50.8 & 51.5 & 50.3 & 47.6
    \\ \midrule
        \multicolumn{10}{l}{\textit{HyperTuning}} \\
    \ \ HyperT5-Prefix
    & 38.7 & 33.6 & 69.6 & 88.4 & 79.5 & 53.1 & 57.6 & 56.6 & 59.6
    \\
    \ \ HyperT5-LoRA
    & 35.3 & 30.8 & 66.4 & 83.3 & 68.5 & 50.3 & 60.0 & 56.1 & 56.4
    \\ \midrule
        \multicolumn{10}{l}{\textit{Other results}} \\
    \ \ T0
    & 33.4 & 27.3 & 45.4 & 73.1 & 64.5 & 50.7 & 65.0 & 51.0 & 51.3
    \\ \bottomrule
\end{tabular}%
\caption{
  Results on P3 on held-out tasks (dev) with T5-XL models.
  T0 results taken from \citet{sanh2022t0}.
}
\label{tab:table_02_p3_3b}
\end{table}


Table~\ref{tab:table_01_p3} and Table~\ref{tab:table_02_p3_3b} show the results of our experiments on the P3 dataset using T5-Large ($\sim$770M parameters) and T5-XL ($\sim$3B parameters), respectively.

We compare our HyperT5-Prefix and HyperT5-LoRA, which use hypermodels to generate task-specific PEFT parameters based on few-shot examples, with several baselines: prefix tuning, LoRA tuning, T5-MTF, and T5-MTF-Few-shot. 
T5-MTF is a model that roughly corresponds to the T0 model, and we detail the differences in Appendix~\ref{app:p3details}.


Our results show that both HyperT5-Prefix and HyperT5-LoRA significantly improve over the prefix and LoRA tuning baselines, indicating the effectiveness of using hypermodels to adapt the frozen downstream T5 model to unseen tasks. 
HyperT5-Prefix achieves performance close to T5-MTF, while T5-MTF-Few-shot attains the highest scores, in line with the findings of \citet{min-etal-2022-metaicl}.
These patterns are consistent across T5-Large and T5-XL,\footnote{We note that T0-XL performs much worse than our trained T5-MTF, which is in agreement with other work \citep{anonymous2023metrot5,wu-etal-2022-continued} that have reported similar results in replicating T0.} demonstrating the scalability of hypertuning.


We emphasize that HyperT5-Prefix/LoRA only introduces a very small number of PEFT parameters in the frozen downstream T5 model, whereas all parameters are tuned in the T5-MTF and T5-MTF-Few-shot models.
Moreover, the P3 examples are written with prompt templates that are optimized for zero-shot inference, which is the ideal input format for T5-MTF. 
Furthermore, T5-MTF-Fewshot has full, bidirectional self-attention between the target input $x$ and the few-shot examples, whereas HyperT5-Prefix and HyperT5-Lora only incorporate information from the few-shot examples via the respective PEFT parameters.


To investigate whether the hypermodel benefits are complementary to updating the downstream model parameters, we conduct an additional set of experiments where we jointly train both the hypermodel and the downstream model (HyperTuning + Fine-Tuning), with results shown at the bottom of Table~\ref{tab:table_01_p3}. 
We observe that both HyperT5-Prefix+ and HyperT5-Lora+ slightly surpass T5-MTF-Fewshot, suggesting that the hypermodels can further enhance the performance of fine-tuned downstream models.

\subsubsection{MetaICL}

\label{sec:results_metaicl}


Table~\ref{tab:table_03_metaicl} presents the results on three MetaICL task splits. 
As in the previous experiments, both HyperT5 models surpass the PEFT models and T5-MTF in performance, except for T5-MTF-Few-shot, which outperforms them in all but one case: Non-NLI$\rightarrow$NLI, where HyperT5-Prefix achieves a higher score.
T5-MTF performs poorly in the MetaICL experiments, as it has to handle task examples zero-shot, and the MetaICL inputs are not suitable for zero-shot inference, as explained above.

\begin{table}[th]
\centering
\small
\begin{tabular}{lrrrrrrrrrrr}
    \toprule
    & \multirow{2}{*}{\shortstack{HR\\$\rightarrow$LR}}
    & \multirow{2}{*}{\shortstack{Non-NLI\\$\rightarrow$NLI}} 
    & \multirow{2}{*}{\shortstack{Non-Class\\$\rightarrow$Class}}
    & 
    \\ &&&
    \\ \midrule
    \multicolumn{4}{l}{\textit{Full Fine-Tuning}} \\
    \ \ T5-MTF
      & 34.3 & 48.8 & 30.3
    \\
    \ \ T5-MTF-Few-shot
      & 41.0 & 56.7 & 40.6
    \\ \midrule
    \multicolumn{4}{l}{\textit{Parameter-Efficient Fine-Tuning (PEFT)}} \\
    \ \ T5-MTF (Prefix)
      & 29.8 & 42.8 & 29.6
    \\
    \ \ T5-MTF (LoRA)
      & 31.5 & 41.3 & 28.7 
    \\ \midrule
      \multicolumn{4}{l}{\textit{HyperTuning}} \\
    \ \ HyperT5-Prefix
      & 38.0 & 58.3 & 38.6
    \\
    \ \ HyperT5-LoRA
      & 35.4 & 54.2 & 34.8
    \\ \bottomrule
\end{tabular}%
\caption{
  Results on MetaICL (Test) with T5-Large models.
}
\label{tab:table_03_metaicl}
\end{table}

\subsubsection{Super-NaturalInstructions (S-NI)}


We report the results on the different S-NI settings in Table~\ref{tab:table_04_natinst} for T5-Large and Table~\ref{tab:table_05_natinst_3b} for T5-XL, using both Def (definition-only) and Def+2Pos (definition and two fixed positive examples) settings.
The T5-MTF (Def) and T5-MTF (Def+2Pos) models are similar to the corresponding T$k$-Instruct variants \citep{wang2022sni}, with a slight difference in input formatting (see Appendix~\ref{app:inputformatting}).
For the hypermodels, we prepend the task definitions to the few-shot examples and treat them as part of the hypermodel input.
On average, the HyperT5 with Def+2Pos outperforms T5-MTF (Def) by a large margin, but still underperforms T5-MTF (Def+2Pos), in line with the above results.

\begin{table}[th]
\parbox{.45\linewidth}{
    \centering
    \small
    \begin{tabular}{lrrrrrrrrrrr}
        \toprule
         & AVG
    \\ \midrule
    \multicolumn{2}{l}{\textit{Full Fine-Tuning}} \\
    \ \ T5-MTF (Def) & 40.6
        \\
    \ \ T5-MTF (Def+2Pos) & 47.6
    \\ \midrule
    \multicolumn{2}{l}{\textit{HyperTuning}} \\
    \ \ HyperT5-Prefix (Def) & 37.1
        \\
    \ \ HyperT5-Prefix (Def+2Pos) & 43.5
        \\
    \ \ HyperT5-LoRA (Def) & 34.9
        \\
    \ \ HyperT5-LoRA (Def+2Pos) & 42.0
        \\
    \midrule
    \multicolumn{2}{l}{\textit{Other Results}} \\
    \ \ T$k$-Instruct (Def+2Pos) & 48.0
    \\
        \bottomrule
    \end{tabular}%
    \caption{
        Results on Super-NaturalInstuctions (S-NI; Test) with T5-Large models.
        T$k$-Instruct results taken from \citet{wang2022sni}.
    }
    \label{tab:table_04_natinst}
}
\parbox{.07\linewidth}{\ }
\parbox{.45\linewidth}{
    \centering
    \small
    \begin{tabular}{lrrrrrrrrrrr}
        \toprule
         & AVG
    \\ \midrule
    \multicolumn{2}{l}{\textit{Full Fine-Tuning}} \\
    \ \ T5-MTF (Def) & 46.6
        \\
    \ \ T5-MTF (Def+2Pos) & 54.3
    \\ \midrule
    \multicolumn{2}{l}{\textit{HyperTuning}} \\
    \ \ HyperT5-Prefix (Def) & 38.9
        \\
    \ \ HyperT5-Prefix (Def+2Pos) & 48.6
        \\
    \ \ HyperT5-LoRA (Def) & 38.9
        \\
    \ \ HyperT5-LoRA (Def+2Pos) & 45.0
        \\
    \midrule
    \multicolumn{2}{l}{\textit{Other Results}} \\
    \ \ T$k$-Instruct (Def+2Pos) & 54.0
    \\
        \bottomrule
    \end{tabular}%
    \caption{
        Results on Super-NaturalInstuctions (S-NI; Test) with T5-XL models.
        T$k$-Instruct results taken from \citet{wang2022sni}.
    }
    \label{tab:table_05_natinst_3b}
}
\end{table}

\subsection{Discussion}


Above, we evaluated hypermodels on three multi-task datasets, where they generate task-specific soft prefixes or LoRA parameters from a few examples or instructions. 
In general, HyperT5 matched or exceeded T5-MTF models, but lagged behind T5-MTF-Fewshot models (or Def+2Pos models, in the case of S-NI). 
This gap is expected, as T5-MTF-Fewshot uses full self-attention between the examples and the target input $x$, while HyperT5 encodes the examples into PEFT parameters that are independent of $x$.
We attribute some of the gap to this limitation.


However, this limitation also confers efficiency advantages to HyperT5 at inference time compared to T5-MTF-Fewshot.
In encoder-decoders such as T5, the full self-attention between the examples and $x$ prevents the separation of their representations: a new forward pass is needed for each new $x$. 
In contrast, for hypermodels the examples can be encoded into PEFT parameters once, and reused for all subsequent inputs.
Even for decoder-only models (e.g. MetaICL based on GPT-2), where the examples can be cached as key and value representations, the cache size is likely much larger than the PEFT parameters, as the cache stores all the representations for every token in the examples, which are several times longer than the input by definition.
Thus, hypermodels in our setup sacrifice some performance for efficiency.


Regarding T5-MTF, one might wonder what the concrete benefit of HyperT5 is, given their similar performance. 
After all, unlike T5-MTF-Fewshot, T5-MTF only uses $x$ as the input, requiring no extra computation or memory, and only one set of model weights.
Firstly, we stress that the HyperT5 model can only affect the downstream model through a small number of modified parameters, while in T5-MTF all the parameters that process $x$ are modified.
Although HyperT5 and T5-MTF have roughly the same number of tuned parameters, the parameters modified in T5-MTF directly interact with the input $x$, which we expect to help performance.
Secondly, we identify two separate, but possibly related, sources of performance improvement: better general task performance of the downstream model (which is usually the goal of MTF training), and adapting the downstream model to a new task based on few-shot examples, using hypermodels in our case. 
Our aim in this work is to show the feasibility of the latter. We argue that both sources are complementary, and we showed in Section~\ref{sec:results_p3} that when we use hypermodels without freezing the downstream model, thereby acquiring both benefits, performance further improves. 
More generally, we expect that training a hypermodel against an already multi-task fine-tuned model will lead to better performance than just using the model for zero-shot inference alone, and we plan to explore this in future work.

We also observe a consistent trend where HyperT5-Prefix outperforms HyperT5-LoRA.
We speculate that it is easier for hypermodels to learn to generate soft prefixes as compared to LoRA weights, since soft prefix are effectively model-internal hidden states, and the generated PEFT parameters are themselves transformations of the hypermodel hidden states.
Incidentally, another possible interpretation of the HyperT5-Prefix model is that the combination of the hypermodel and the downstream model can be seen as a dual-encoder, single-decoder model with separate encoders for the few-shot examples and the target example.

Lastly, the majority of the experiments were conducted with minimal hyperparameter-tuning, and the current results primarily serve as a proof-of-concept of hypertuning being a viable approach to adapt downstream models.
We expect that further exploration of hyperpretraining and MTF hyperparameters as well as hypermodel architectures may lead to better results and overcome some of the limitations we identified.

\label{sec:results_natinst}

\subsection{Is HyperPretraining Necessary?}
\label{sec:hyperpretrainingresults}


\begin{figure}
\begin{subfigure}{.5\textwidth}
  \centering
  \includegraphics[width=1.0\linewidth]{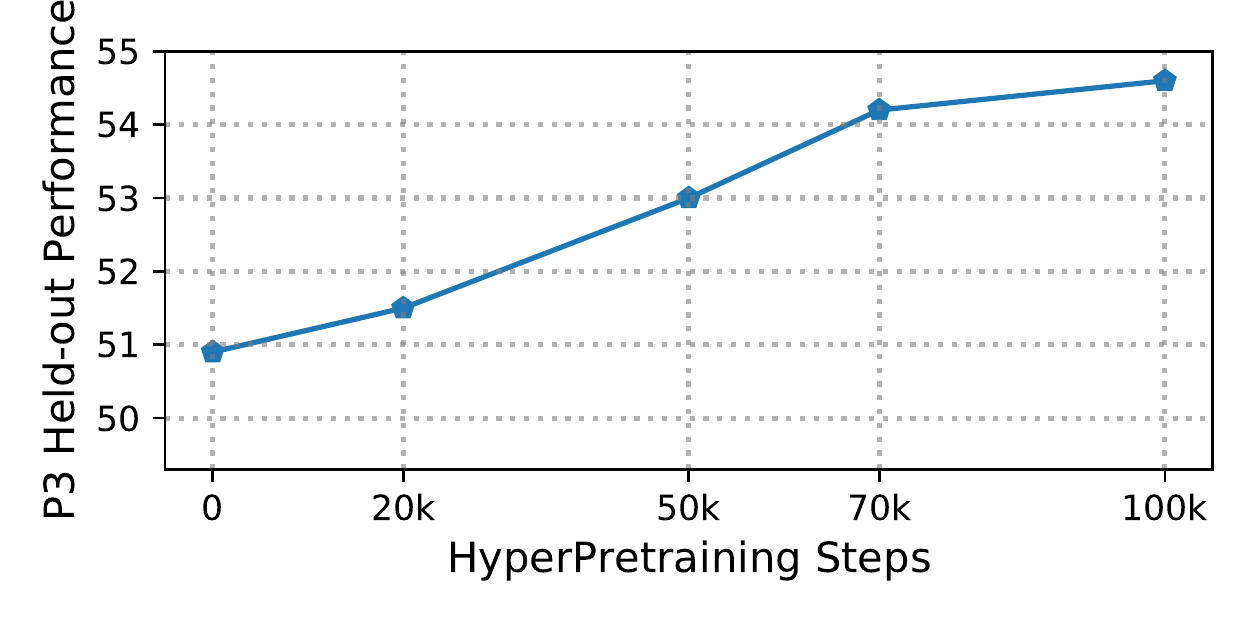}
  \caption{HyperT5-Prefix}
\end{subfigure}%
\begin{subfigure}{.5\textwidth}
  \centering
  \includegraphics[width=1.0\linewidth]{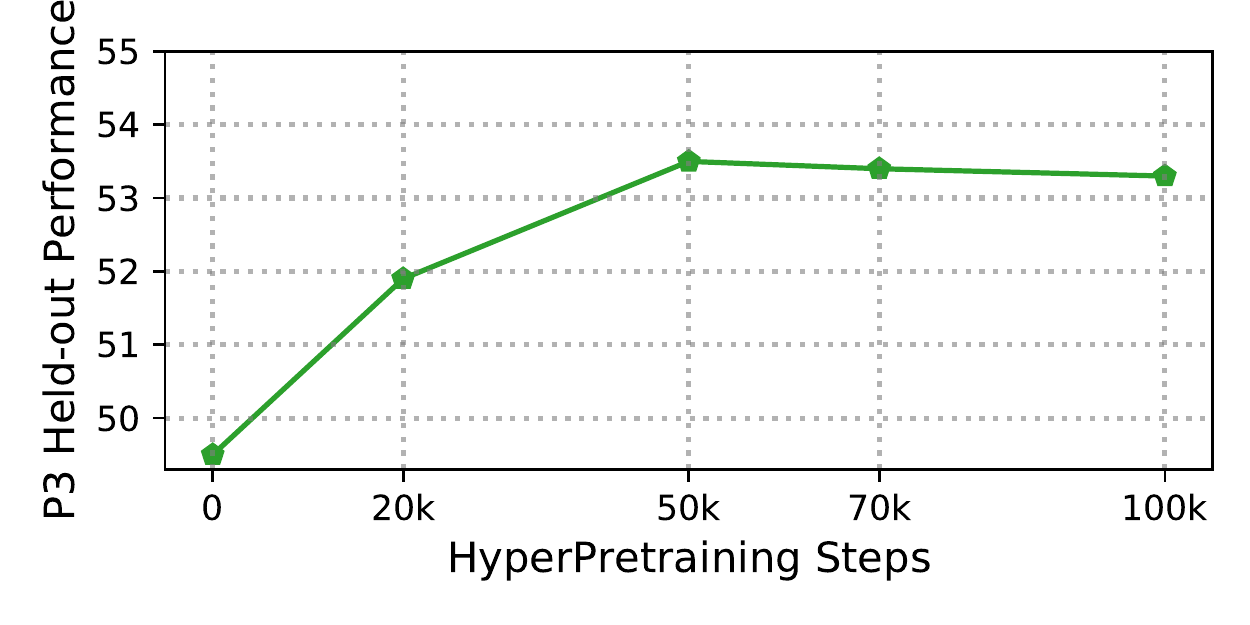}
  \caption{HyperT5-LoRA}
\end{subfigure}
\caption{
  Performance of HyperT5 models on P3 evaluation with different amounts of hyperpretraining.
  HyperPretraining is crucial for good performance of the hypermodels.
  However, hyperpretraining for too many steps can also hurt performance (as see in the case of HyperT5-LoRA).
}
\label{fig:hyperpretraining_steps}
\end{figure}


We demonstrate the benefits of hyperpretraining for the hypermodels in this section.
As mentioned in Section~\ref{fig:hyperpretraining}, we hyperpretrained the hypermodels for 100k steps before multi-task fine-tuning them on P3 tasks.
To examine the impact of hyperpretraining, we also multi-task fine-tuned HyperT5-Prefix and HyperT5-LoRA from LM-adapted T5 without any hyperpretraining, and from intermediate checkpoints over the course of hyperpretraining.
Figure~\ref{fig:hyperpretraining_steps} shows the average scores on the held-out tasks for these models.
Both HyperT5 models perform very poorly without any hyperpretraining, achieving scores similar to PEFT-only (see Table~\ref{tab:table_01_p3}).
With hyperpretraining, the performance of both hypermodels significantly improves.
While HyperT5-Prefix appears to consistently improve over the course of 100k steps, we observe that HyperT5-LoRA performance slightly declines after 50k steps.
Hypermodels targeting different PEFT methods may benefit from different amounts of hyperpretraining, and we emphasize that our choice of the number of hyperpretraining steps is by no means considered to be optimal.\footnote{We chose 100k steps based on the T5 LM-adaptation procedure \citep{lester2021prompt}. } 
We expect that better hyperpretraining configurations can be explored in future work.

\section{HyperModels for Improved Parameter Initialization}

\label{sec:initialization}

Thus far, we have discussed hypermodels in the context of generating PEFT parameters in a single forward pass through the hypermodel.
We can also consider an alternative use of hypermodels: Instead of randomly initializing new parameters, we can use hypermodels to produce task-specific PEFT parameters based on a few examples from the task.
This can be seen as using task knowledge acquired by the hypermodel during training to provide a first approximation of PEFT parameters, and thereafter refining the parmaeters via regular PEFT training.

In conventional PEFT, wherever new parameters are introduced into the model, they are either initialized randomly, or with fixed initial values (e.g. the up-projection weights in LoRA are initialized to 0)--for brevity, we will refer to this simply as random initialization.
Beyond random initialization, \citet[][SPoT]{vu2021spot} and \citet[][PPT]{gu2021ppt} have explored transfer-learning within PEFT, first doing PEFT on one or more upstream tasks, and then using the learned PEFT parameters as an initialization for downstream PEFT.

This approach has two advantages over conventional PEFT initializations.
First, the hypermodel-generated parameters already perform well on the task, as shown in Section~\ref{sec:results}, so PEFT training can reach good performance faster.
Second, the hypermodel can automatically transfer relevant knowledge from previous tasks to the new task, similar to SPoT and PPT, except we let the hypermodel determine what previously learned task knowledge is most applicable to the new task.
For instance, a major challenge addressed in SPoT was searching for the set of upstream tasks whose PEFT parameters would be the most appropriate initialization for a downstream task--in our case, we can directly provide a hypermodel with few-shot examples to generate our desired initialization.

To investigate the effectiveness of using hypermodels to generate PEFT initializations, we use the P3-trained models from Section~\ref{sec:results_p3}, and perform prefix tuning and LoRA tuning on the held-out tasks individually.\footnote{We use one specific prompt format for each task, listed in Appendix~\ref{app:p3details}.}
For each method-task pair, we sweep across learning rates $\{1e^{-3}, 1e^{-4}, 1e^{-5}\}$ and take the best average result over 3 random seeds.

We consider two baselines for initializations: random initialization (Rand Init) 
and using the multi-task fine-tuned PEFT parameters from Section~\ref{sec:results_p3} as initializations (Shared Init).
The hypermodel-generated initialization (Hyper Init) is generated using a randomly sampled set of 16 examples from the respective training sets.

\begin{figure}
\begin{subfigure}{.5\textwidth}
  \centering
  \includegraphics[width=1\linewidth]{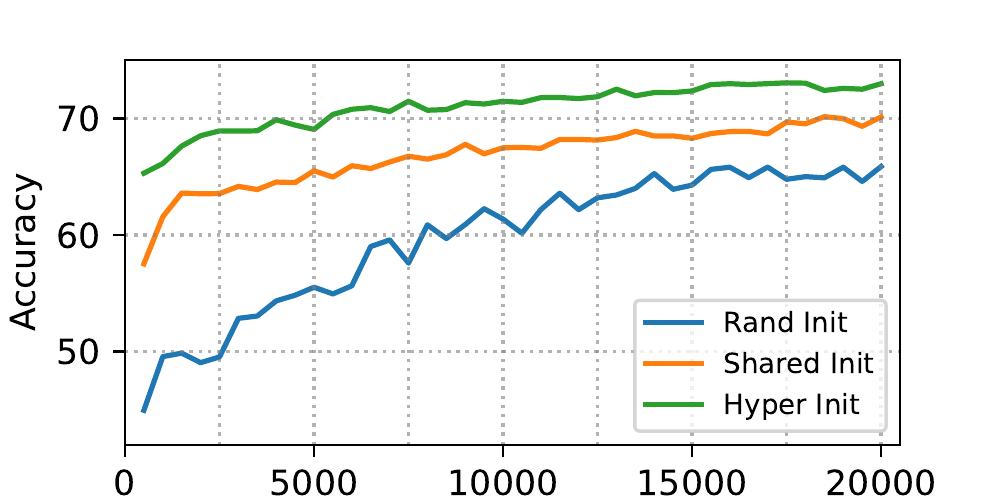}
  \caption{Prefix Tuning}
\end{subfigure}%
\begin{subfigure}{.5\textwidth}
  \centering
  \includegraphics[width=1\linewidth]{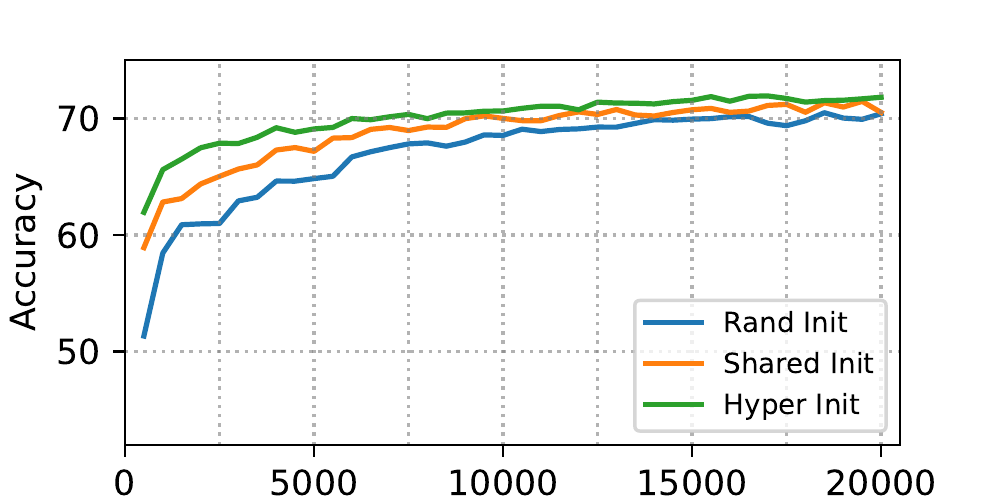}
  \caption{LoRA}
\end{subfigure}
\caption{
  Average performance on P3 held-out tasks with prefix tuning and LoRA, using different parameter initializations.
  Using hypermodel-generated initializations starts with higher performance and continues to perform better on average over the course of training.
}
\label{fig:peft_init}
\end{figure}

We show the results of prefix tuning\footnote{Prefix tuning is performed via a reparameterization, in line with standard practice. Refer to Appendix~\ref{app:prefixtuning} for details.} and LoRA tuning with different initialization schemes in Table~\ref{tab:table_07_peft}. 
We observe that for both prefix tuning and LoRA tuning, shared initialization significantly performs random initialization, while using a hypermodel-generated initialization outperforms both on average.
We also show the average performance across tasks over the course of tuning in Figure~\ref{fig:peft_init}. 
We observe that hypermodel-generated initializations start with much better performance compared to the other two initialization schemes, and continue to outperform them over the course of fine-tuning.
Hence, hypermodels can be complementary to a standard PEFT pipeline, providing both performance gains and computational cost savings.

\begin{table}[th]
\centering
\small
\begin{tabular}{lrrrrrrrrrrr}
    \toprule
    & ANLI & HSwg & CB & COPA & RTE & WiC & WSC & WGD & \underline{AVG}
    \\ \midrule
    Prefix (Rand Init)
  & 54.6 & 50.5 & 98.8 & 79.0 & 78.8 & 71.6 & 63.5 & 52.2 & 68.6
    \\
    Prefix (Shared Init)
  & 60.8 & 51.6 & 99.4 & 85.7 & 84.8 & 72.4 & 72.6 & 65.1 & 74.0
    \\
    Prefix (Hyper Init)
  & 61.4 & 51.5 & 97.6 & 84.3 & 87.1 & 71.2 & 76.5 & 71.6 & 75.2
    \\ \midrule
    LoRA (Rand Init)
  & 59.5 & 51.3 & 93.5 & 78.0 & 82.6 & 73.5 & 77.9 & 65.1 & 72.7
    \\
    LoRA (Shared Init)
  & 57.9 & 51.6 & 99.4 & 83.0 & 83.8 & 73.1 & 73.3 & 67.9 & 73.7
    \\
    LoRA (Hyper Init)
  & 57.7 & 48.4 & 99.4 & 87.3 & 84.1 & 73.0 & 83.9 & 66.2 & 75.0
    \\ \bottomrule
\end{tabular}%
\caption{
  Prefix tuning and LoRA fine-tuning on T5-Large with different initializations on P3 held-out tasks.
  Using HyperT5-generated parameters as an initialization achieves better performance on average than using shared MTF PEFT parameters or random initialization.
}
\label{tab:table_07_peft}
\end{table}

\section{Conclusion}


We introduce the concept of \textit{hypertuning}, which leverages a hypermodel to adapt a downstream model to a specific downstream application.
We present a basic framework for hypertuning, where a hypermodel is trained to produce parameters for a downstream model from few-shot examples in one forward pass, and we apply this framework to train HyperT5-Prefix and HyperT5-LoRA models that can adapt a fixed downstream T5 model.
We find that a two-stage training procedure of hyperpretraining and multi-task fine-tuning is effective for training hypermodels, and we evaluate the HyperT5 models on P3, MetaICL and S-NI datasets, showing that they can generate PEFT parameters that enable the downstream T5 models to perform well on unseen tasks.
Furthermore, the parameters generated by hypertuning can also serve as improved parameter initializations for parameter-efficient fine-tuning.
We regard these findings as an initial but encouraging indication of the potential of adapting large language models without back-propagation.

\section{Acknowledgements}

We would like to thank Sam Bowman for their thoughtful feedback and Jonas Pfeiffer for early idea discussion.

\bibliography{anthology,custom}
\bibliographystyle{acl_natbib}

\clearpage
\appendix

\section{Training Details}
\label{app:trainingdetails}
All experiments are trained with 1-bit Adam \citep{dettmers2022optimizers} and batch size of 256, a learning rate of 5e-5, and a linear decay schedule.
Training was performed with ZeRO \citep{rajbhandari2020zero} and Transformers \citep{wolf-etal-2020-transformers}.
For hypermodels, the hypermodel's max input sequence length is 1024 tokens and the downstream model's max input sequence length is 384 tokens.
Correspondingly, the max input sequence length for all non-few-shot models (e.g. T5-MTF, T5-MTF(Prefix)) is 384.
The max input sequence length of few-shot models (e.g. T5-MTF-Few-shot) is thus conservatively set at 1024+384=1408 tokens.
The max target sequence length is set to 128 for all experiments.

\subsection{Input Formatting}
\label{app:inputformatting}

Few-shot examples for hypermodels are formatted in the following manner:

\begin{zitat}{}
\texttt{
\scriptsize
    <x> Input 1 <y> Target 1
    <x> Input 2 <y> Target 2
    <x> Input 3 <y> Target 3
}
\end{zitat}

where \texttt{<x>} and \texttt{<y>} and special tokens.

For S-NI, the task definitions are treated as just another example: 

\begin{zitat}{}
\texttt{
\scriptsize
    <x> Instruction
    <x> Input 1 <y>Target 1
    <x> Input 2 <y>Target 2
}
\end{zitat}

\section{Dataset-specific Details}

\label{app:datadetails}

\subsection{P3 / T0}

\label{app:p3details}

We highlight some differences our T0 baselines and the T0 setup described in the original paper \citep{sanh2022t0}.
Besides the different optimizers and batch sizes listed above, we do not use packing to process our training data.
Moreover, because our focus is on few-shot learning, we remove a number of tasks formulations with longer inputs from the T0-train dataset, listed in Section~\ref{fig:appp3train}.
For T0, we use an input sequence length of 384 and output length of 128, which matches the input and output lengths of the downstream model in our hypermodel setup.
For T5-MTF-Few-shot, we use an input sequence length of 1024+384=1408, which is the combined input lengths of the hypermodel and downstream model.
We believe that these changes can meaningfully modify the performance of the T0 models, but provide a fairer baseline to the hypermodel setup.

\begin{figure}
\begin{minipage}[t]{\linewidth}\raggedright
\tiny{adversarial\_qa\_dbert\_answer\_the\_following\_q, adversarial\_qa\_dbert\_based\_on, adversarial\_qa\_dbert\_generate\_question, adversarial\_qa\_dbert\_question\_context\_answer, adversarial\_qa\_dbert\_tell\_what\_it\_is, adversarial\_qa\_dbidaf\_answer\_the\_following\_q, adversarial\_qa\_dbidaf\_based\_on, adversarial\_qa\_dbidaf\_generate\_question, adversarial\_qa\_dbidaf\_question\_context\_answer, adversarial\_qa\_dbidaf\_tell\_what\_it\_is, adversarial\_qa\_droberta\_answer\_the\_following\_q, adversarial\_qa\_droberta\_based\_on, adversarial\_qa\_droberta\_generate\_question, adversarial\_qa\_droberta\_question\_context\_answer, adversarial\_qa\_droberta\_tell\_what\_it\_is, ag\_news\_classify, ag\_news\_classify\_question\_first, ag\_news\_classify\_with\_choices, ag\_news\_classify\_with\_choices\_question\_first, ag\_news\_recommend, ag\_news\_which\_section, ag\_news\_which\_section\_choices, amazon\_polarity\_Is\_this\_product\_review\_positive, amazon\_polarity\_Is\_this\_review, amazon\_polarity\_Is\_this\_review\_negative, amazon\_polarity\_User\_recommend\_this\_product, amazon\_polarity\_convey\_negative\_or\_positive\_sentiment, amazon\_polarity\_flattering\_or\_not, amazon\_polarity\_negative\_or\_positive\_tone, amazon\_polarity\_user\_satisfied, amazon\_polarity\_would\_you\_buy, app\_reviews\_categorize\_rating\_using\_review, app\_reviews\_convert\_to\_rating, app\_reviews\_convert\_to\_star\_rating, app\_reviews\_generate\_review, cnn\_dailymail\_3.0.0\_generate\_story, cnn\_dailymail\_3.0.0\_spice\_up\_story, common\_gen\_Example\_prompt, common\_gen\_Given\_concepts\_type\_1, common\_gen\_Given\_concepts\_type\_2, common\_gen\_Put\_together, common\_gen\_choice\_in\_concept\_centric\_sentence\_generation, common\_gen\_random\_task\_template\_prompt, common\_gen\_sentence\_to\_concepts, common\_gen\_topic\_to\_sentence, common\_gen\_topics\_from\_the\_sentence, cos\_e\_v1.11\_aligned\_with\_common\_sense, cos\_e\_v1.11\_description\_question\_option\_id, cos\_e\_v1.11\_description\_question\_option\_text, cos\_e\_v1.11\_explain\_why\_human, cos\_e\_v1.11\_generate\_explanation\_given\_text, cos\_e\_v1.11\_i\_think, cos\_e\_v1.11\_question\_description\_option\_id, cos\_e\_v1.11\_question\_description\_option\_text, cos\_e\_v1.11\_question\_option\_description\_id, cos\_e\_v1.11\_question\_option\_description\_text, cos\_e\_v1.11\_rationale, cosmos\_qa\_context\_answer\_to\_question, cosmos\_qa\_context\_description\_question\_answer\_id, cosmos\_qa\_context\_description\_question\_answer\_text, cosmos\_qa\_context\_description\_question\_text, cosmos\_qa\_context\_question\_description\_answer\_id, cosmos\_qa\_context\_question\_description\_answer\_text, cosmos\_qa\_context\_question\_description\_text, cosmos\_qa\_description\_context\_question\_answer\_id, cosmos\_qa\_description\_context\_question\_answer\_text, cosmos\_qa\_description\_context\_question\_text, cosmos\_qa\_no\_prompt\_id, cosmos\_qa\_no\_prompt\_text, cosmos\_qa\_only\_question\_answer, dbpedia\_14\_given\_a\_choice\_of\_categories\_, dbpedia\_14\_given\_a\_list\_of\_category\_what\_does\_the\_title\_belong\_to, dbpedia\_14\_given\_list\_what\_category\_does\_the\_paragraph\_belong\_to, dbpedia\_14\_pick\_one\_category\_for\_the\_following\_text, dream\_answer\_to\_dialogue, dream\_baseline, dream\_generate\_first\_utterance, dream\_generate\_last\_utterance, dream\_read\_the\_following\_conversation\_and\_answer\_the\_question, duorc\_ParaphraseRC\_build\_story\_around\_qa, duorc\_SelfRC\_build\_story\_around\_qa, gigaword\_TLDR, gigaword\_first\_sentence\_title, gigaword\_generate\_summary\_for\_this, gigaword\_in\_a\_nutshell, gigaword\_make\_a\_title, gigaword\_reverse\_writing, gigaword\_write\_a\_title\_for\_this\_sentence, gigaword\_write\_an\_article, gigaword\_write\_its\_sentence, glue\_mrpc\_equivalent, glue\_mrpc\_generate\_paraphrase, glue\_mrpc\_generate\_sentence, glue\_mrpc\_paraphrase, glue\_mrpc\_replace, glue\_mrpc\_same\_thing, glue\_mrpc\_want\_to\_know, glue\_qqp\_answer, glue\_qqp\_duplicate, glue\_qqp\_duplicate\_or\_not, glue\_qqp\_meaning, glue\_qqp\_quora, glue\_qqp\_same\_thing, imdb\_Movie\_Expressed\_Sentiment, imdb\_Movie\_Expressed\_Sentiment\_2, imdb\_Negation\_template\_for\_positive\_and\_negative, imdb\_Reviewer\_Enjoyment, imdb\_Reviewer\_Enjoyment\_Yes\_No, imdb\_Reviewer\_Expressed\_Sentiment, imdb\_Reviewer\_Opinion\_bad\_good\_choices, imdb\_Reviewer\_Sentiment\_Feeling, imdb\_Sentiment\_with\_choices\_, imdb\_Text\_Expressed\_Sentiment, imdb\_Writer\_Expressed\_Sentiment, kilt\_tasks\_hotpotqa\_combining\_facts, kilt\_tasks\_hotpotqa\_complex\_question, kilt\_tasks\_hotpotqa\_final\_exam, kilt\_tasks\_hotpotqa\_formulate, kilt\_tasks\_hotpotqa\_straighforward\_qa, paws\_labeled\_final\_Concatenation, paws\_labeled\_final\_Concatenation\_no\_label, paws\_labeled\_final\_Meaning, paws\_labeled\_final\_Meaning\_no\_label, paws\_labeled\_final\_PAWS\_ANLI\_GPT3, paws\_labeled\_final\_PAWS\_ANLI\_GPT3\_no\_label, paws\_labeled\_final\_Rewrite, paws\_labeled\_final\_Rewrite\_no\_label, paws\_labeled\_final\_context\_question, paws\_labeled\_final\_context\_question\_no\_label, paws\_labeled\_final\_paraphrase\_task, paws\_labeled\_final\_task\_description\_no\_label, qasc\_is\_correct\_1, qasc\_is\_correct\_2, qasc\_qa\_with\_combined\_facts\_1, qasc\_qa\_with\_separated\_facts\_1, qasc\_qa\_with\_separated\_facts\_2, qasc\_qa\_with\_separated\_facts\_3, qasc\_qa\_with\_separated\_facts\_4, qasc\_qa\_with\_separated\_facts\_5, quarel\_choose\_between, quarel\_do\_not\_use, quarel\_heres\_a\_story, quarel\_logic\_test, quarel\_testing\_students, quartz\_answer\_question\_based\_on, quartz\_answer\_question\_below, quartz\_given\_the\_fact\_answer\_the\_q, quartz\_having\_read\_above\_passage, quartz\_paragraph\_question\_plain\_concat, quartz\_read\_passage\_below\_choose, quartz\_use\_info\_from\_paragraph\_question, quartz\_use\_info\_from\_question\_paragraph, ropes\_background\_new\_situation\_answer, ropes\_background\_situation\_middle, ropes\_given\_background\_situation, ropes\_new\_situation\_background\_answer, ropes\_plain\_background\_situation, ropes\_plain\_bottom\_hint, ropes\_plain\_no\_background, ropes\_prompt\_beginning, ropes\_prompt\_bottom\_hint\_beginning, ropes\_prompt\_bottom\_no\_hint, ropes\_prompt\_mix, ropes\_read\_background\_situation, rotten\_tomatoes\_Movie\_Expressed\_Sentiment, rotten\_tomatoes\_Movie\_Expressed\_Sentiment\_2, rotten\_tomatoes\_Reviewer\_Enjoyment, rotten\_tomatoes\_Reviewer\_Enjoyment\_Yes\_No, rotten\_tomatoes\_Reviewer\_Expressed\_Sentiment, rotten\_tomatoes\_Reviewer\_Opinion\_bad\_good\_choices, rotten\_tomatoes\_Reviewer\_Sentiment\_Feeling, rotten\_tomatoes\_Sentiment\_with\_choices\_, rotten\_tomatoes\_Text\_Expressed\_Sentiment, rotten\_tomatoes\_Writer\_Expressed\_Sentiment, samsum\_Generate\_a\_summary\_for\_this\_dialogue, samsum\_Given\_the\_above\_dialogue\_write\_a\_summary, samsum\_Sum\_up\_the\_following\_dialogue, samsum\_Summarize\_, samsum\_Summarize\_this\_dialogue\_, samsum\_To\_sum\_up\_this\_dialog, samsum\_Write\_a\_dialogue\_that\_match\_this\_summary, sciq\_Direct\_Question, sciq\_Direct\_Question\_Closed\_Book\_, sciq\_Multiple\_Choice, sciq\_Multiple\_Choice\_Closed\_Book\_, sciq\_Multiple\_Choice\_Question\_First, social\_i\_qa\_Check\_if\_a\_random\_answer\_is\_valid\_or\_not, social\_i\_qa\_Generate\_answer, social\_i\_qa\_Generate\_the\_question\_from\_the\_answer, social\_i\_qa\_I\_was\_wondering, social\_i\_qa\_Show\_choices\_and\_generate\_answer, social\_i\_qa\_Show\_choices\_and\_generate\_index, trec\_fine\_grained\_ABBR, trec\_fine\_grained\_ABBR\_context\_first, trec\_fine\_grained\_DESC, trec\_fine\_grained\_DESC\_context\_first, trec\_fine\_grained\_ENTY, trec\_fine\_grained\_HUM, trec\_fine\_grained\_HUM\_context\_first, trec\_fine\_grained\_LOC, trec\_fine\_grained\_LOC\_context\_first, trec\_fine\_grained\_NUM, trec\_fine\_grained\_NUM\_context\_first, trec\_fine\_grained\_open, trec\_fine\_grained\_open\_context\_first, trec\_pick\_the\_best\_descriptor, trec\_trec1, trec\_trec2, trec\_what\_category\_best\_describe, trec\_which\_category\_best\_describes, wiki\_bio\_comprehension, wiki\_bio\_guess\_person, wiki\_bio\_key\_content, wiki\_bio\_what\_content, wiki\_bio\_who, wiki\_qa\_Decide\_good\_answer, wiki\_qa\_Direct\_Answer\_to\_Question, wiki\_qa\_Generate\_Question\_from\_Topic, wiki\_qa\_Is\_This\_True\_, wiki\_qa\_Jeopardy\_style, wiki\_qa\_Topic\_Prediction\_Answer\_Only, wiki\_qa\_Topic\_Prediction\_Question\_Only, wiki\_qa\_Topic\_Prediction\_Question\_and\_Answer\_Pair, wiki\_qa\_automatic\_system, wiki\_qa\_exercise, wiki\_qa\_found\_on\_google, wiqa\_does\_the\_supposed\_perturbation\_have\_an\_effect, wiqa\_effect\_with\_label\_answer, wiqa\_effect\_with\_string\_answer, wiqa\_what\_is\_the\_final\_step\_of\_the\_following\_process, wiqa\_what\_is\_the\_missing\_first\_step, wiqa\_what\_might\_be\_the\_first\_step\_of\_the\_process, wiqa\_what\_might\_be\_the\_last\_step\_of\_the\_process, wiqa\_which\_of\_the\_following\_is\_the\_supposed\_perturbation, yelp\_review\_full\_based\_on\_that, yelp\_review\_full\_format\_rating, yelp\_review\_full\_format\_score, yelp\_review\_full\_format\_star, yelp\_review\_full\_on\_a\_scale, yelp\_review\_full\_so\_i\_would, yelp\_review\_full\_this\_place
}
\end{minipage}
\caption{List of P3 dataset-prompts used for training. We chose a subset of T0-train with average input lengths shorter than 320 tokens.}
\label{fig:appp3train}
\end{figure}

For the hypermodel initialization/PEFT experiments, we do single-task parameter-efficient fine-tuning on each of the following dataset-prompts:

\begin{enumerate}
    \item anli\_GPT\_3\_style\_r1
    \item hellaswag\_complete\_first\_then
    \item super\_glue\_cb\_GPT\_3\_style
    \item super\_glue\_copa\_C1\_or\_C2\_premise\_so\_because\_
    \item super\_glue\_rte\_GPT\_3\_style
    \item super\_glue\_wic\_GPT\_3\_prompt
    \item super\_glue\_wsc.fixed\_GPT\_3\_Style
    \item winogrande\_winogrande\_debiased\_Replace

\end{enumerate}

\subsection{S-NI / T-KI}

To standardize the preprocessing across our experiments, we do not use the input formatting provided in the original work \citep{wang2022sni}. 
Instead, we use the format described in Appendix~\ref{app:inputformatting} for all experiments.
Given that the same format is used in multi-task fine-tuning and evaluation, this should not unfairly advantage any model.
However, because the format deviates from that of the original work, we do not directly evaluate the T-KI models.

Additionally, the Super-NaturalInstructions dataset (previously known as NaturalInstructions-v2) has undergone some changes over time.
In our experiments, we use the v2.5 version of the dataset.

\subsection{MetaICL}

\section{Model Details}

\begin{figure}
\begin{lstlisting}
# B = batch_size
# T = input_length
# P = number of prompt tokens
# H = hidden_dim
# L = num layers in encoder/decoder

# Shape: [B, T]
fewshot_input_ids = ...

# Shape: [B, T, H]
hyper_enc_out = hypermodel.encoder(fewshot_input_ids)

# Shape: [B, 2P, H]
# Decoder implicitly uses a fixed set of input embeddings of size 2P
hyper_dec_out = hypermodel.decoder(hyper_enc_out)

# Shape: [B, P, LH]
downstream_enc_k_prefix = hypermodel.enc_k_head(hyper_dec_out[:, :P, :])
downstream_enc_v_prefix = hypermodel.enc_v_head(hyper_dec_out[:, :P, :])
downstream_dec_k_prefix = hypermodel.dec_k_head(hyper_dec_out[:, P:, :])
downstream_dec_v_prefix = hypermodel.dec_v_head(hyper_dec_out[:, P:, :])

# Shape: [B, P, L H]
downstream_enc_k_prefix = downstream_enc_k_prefix.reshape(B, P, L, H)
downstream_enc_v_prefix = downstream_enc_v_prefix.reshape(B, P, L, H)
downstream_dec_k_prefix = downstream_dec_k_prefix.reshape(B, P, L, H)
downstream_dec_v_prefix = downstream_dec_v_prefix.reshape(B, P, L, H)
# These correspond to the per-layer learned prefixes for K and V


# where each of the heads is defined (e.g.):
hypermode.enc_k_head = nn.Sequential([
    nn.LayerNorm(),
    nn.Linear(H),
    nn.TanH(),
    nn.Linear(L*H),
])
\end{lstlisting}
\caption{Pseudo-code for HyperT5-Prefix}
\label{app:pseudoprefix}
\end{figure}

\begin{figure}
\begin{lstlisting}
# B = batch_size
# T = input_length
# R = LoRA rank
# H = hidden_dim
# L = num layers in encoder/decoder

# Shape: [B, T]
fewshot_input_ids = ...

# Shape: [B, T, H]
hyper_enc_out = hypermodel.encoder(fewshot_input_ids)

# Shape: [B, 3L, H]
# Decoder implicitly uses a fixed set of input embeddings of size 3L
hyper_dec_out = hypermodel.decoder(hyper_enc_out)

# Shape: [B, L, H]
enc_repr = hyper_dec_out[:, :L, :]
dec_repr = hyper_dec_out[:, L:2*L, :]
cross_repr = hyper_dec_out[:, 2*L:, :]

# Repeat for dec_repr, cross_repr for decoder self- and cross-attention
# Shape: [B, L, 2RH]
enc_q_repr = hypermodel.enc_q_head(enc_repr)
enc_v_repr = hypermodel.enc_v_head(enc_repr)

# Shape: [B, L, 2RH]
enc_q_repr = enc_q_repr.reshape(B, L, 2, R, H)
enc_v_repr = enc_v_repr.reshape(B, L, 2, R, H)

# raw_enc_q_gate and raw_enc_v_gate are learned parameters of size [L]
# Shape: [1, L, 1, 1, 1]
enc_q_gate = torch.tanh(raw_enc_q_gate)[None, :, None, None, None]
enc_v_gate = torch.tanh(raw_enc_v_gate)[None, :, None, None, None]

# Shape: List of [B, R, H]
enc_lora_q_up_list = [enc_q_repr[:, l, 0, :, :] for l in range(L)]
enc_lora_q_down_list = [enc_q_repr[:, l, 1, :, :] for l in range(L)]
enc_lora_v_up_list = [enc_v_repr[:, l, 0, :, :] for l in range(L)]
enc_lora_v_down_list = [enc_v_repr[:, l, 1, :, :] for l in range(L)]
# These correspond to up- and down-map deltas in LoRA in Q and V 
# attention linear maps


# where each of the heads is defined (e.g.):
hypermode.enc_q_head = nn.Sequential([
    nn.LayerNorm(),
    nn.Linear(H),
    nn.TanH(),
    nn.Linear(2*R*H),
])
\end{lstlisting}
\caption{Pseudo-code for HyperT5-LoRA}
\label{app:pseudolora}
\end{figure}

\section{Elaboration on Prefix Tuning Comparisons}

\label{app:prefixtuning}

While prefix tuning is generally presented as learning a set of prepended key and value representations for each Transformer layer, in practice, the learned prefixes are not optimized directly.
In the work that introduced prefix tuning \citep{li-liang-2021-prefix}, Section~4.3 explains that directly optimizing the learned prefixes leads to unstable training and poorer performance, and instead recommend optimizing a set of learned embeddings and a parameterized MLP to generate the learned prefixes.
(At inference time, the prefixes can be generated from the learned components--this only impacts the training process.)
We confirmed in our experiments that directly optimizing prefixes leads to poor perfomance, and other works involving prefix tuning have similarly used this prefix reparamterization

Hence, we have two flavors of prefix tuning to consider: directly optimizing over prefixes (``Prefix-Flat"), and optimizing with reparamterization (``Prefix-MLP").
The T5-MTF (Prefix) model uses Prefix-MLP, which is the appropriate approach to tuning prefixes in that setting.
However, because HyperT5-Prefix only generates the final prefixes, only Prefix-Flat tuning is possible.
Hence, when we perform the prefix tuning with different initializations in Section~\ref{sec:initialization}, we cannot fairly compare the two methods directly--one which uses a reparameterization during training, and the other which uses direct optimization which we know performs worse in practice.

Instead, we compare prefix tuning in the two different settings, Prefix-Flat and Prefix-MLP, completely separately.
We describe each individual initialization scheme:

\paragraph{Prefix-Flat}

\begin{enumerate}
  \item Prefix-Flat (Rand): Randomly initialize soft prefixes
  \item Prefix-Flat (Shared): Run a forward pass through the prefix reparameterization to obtain the flat prefixes, and use them as the initialization
  \item Prefix-Flat (Hyper): Generate prefixes with HyperT5-Prefix
\end{enumerate}

\paragraph{Prefix-MLP}

\begin{enumerate}
  \item Prefix-MLP (Rand): Randomly initialize the prefix reparameterization embeddings and MLPs (i.e. conventional prefix tuning)
  \item Prefix-MLP (Shared): Directly reuse the prefix reparameterization from T5-MTF (Prefix)
  \item Prefix-MLP (Hyper): We train an entirely new HyperT5-Prefix-MLP model, where the parameter generation heads directly correspond to the prefix tuning reparameterization MLPs. The encoder-decoder in the hypermodel will output the ``embeddings", and we directly reuse the parameter generation heads during tuning.
\end{enumerate}

The results for Prefix-MLP are presented in the body of the paper in Section~\ref{sec:initialization}.
We believe that this approach provides the fairest comparison of initializations.
Importantly, both Prefix-MLP (Shared) and Preflix-MLP (Hyper) have been trained on the same number of labeled examples (not including the few-shot examples, which are inputs), but where the Prefix-MLP uses a single set of learned embeddings, HyperT5-Prefix-MLP generates the embeddings based on few-shot examples.

\begin{figure}
  \centering
  \includegraphics[width=0.5\linewidth]{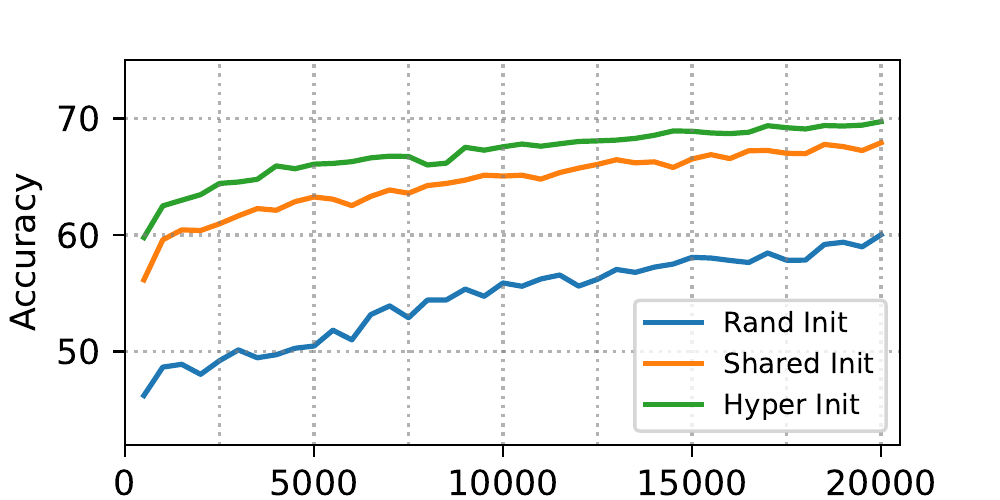}
  \caption{Average performance on P3 held-out tasks with prefix tuning (flat).}
\label{fig:prefix_init}
\end{figure}

We present the full set of prefix tuning results in Table~\ref{tab:table_a01_peft_prefix}, the performance of Prefix-Flat Figure~\ref{fig:prefix_init}.

\begin{table}
\centering
\small
\begin{tabular}{lrrrrrrrrrrr}
    \toprule
    & ANLI & HSwag & CB & COPA & RTE & WiC & WSC & WGD & \underline{AVG}
    \\ \midrule
    Prefix-Flat (Rand Init)
  & 43.6 & 36.3 & 82.7 & 74.0 & 72.9 & 64.4 & 64.2 & 53.0 & 61.4
    \\
    Prefix-Flat (Shared Init)
  & 54.3 & 40.4 & 98.8 & 82.7 & 83.9 & 71.0 & 67.4 & 57.1 & 69.4
    \\
    Prefix-Flat (Hyper Init)
  & 56.6 & 43.5 & 91.7 & 84.3 & 85.3 & 69.3 & 73.0 & 67.6 & 71.4
    \\ \midrule
    Prefix-MLP (Rand Init)
  & 54.6 & 50.5 & 98.8 & 79.0 & 78.8 & 71.6 & 63.5 & 52.2 & 68.6
    \\
    Prefix-MLP (Shared Init)
  & 60.8 & 51.6 & 99.4 & 85.7 & 84.8 & 72.4 & 72.6 & 65.1 & 74.0
    \\
    Prefix-MLP (Hyper Init)
  & 61.4 & 51.5 & 97.6 & 84.3 & 87.1 & 71.2 & 76.5 & 71.6 & 75.2
    \\ \midrule
    LoRA (Rand Init)
  & 59.5 & 51.3 & 93.5 & 78.0 & 82.6 & 73.5 & 77.9 & 65.1 & 72.7
    \\
    LoRA (Shared Init)
  & 57.9 & 51.6 & 99.4 & 83.0 & 83.8 & 73.1 & 73.3 & 67.9 & 73.7
    \\
    LoRA (Hyper Init)
  & 57.7 & 48.4 & 99.4 & 87.3 & 84.1 & 73.0 & 83.9 & 66.2 & 75.0
    \\ \bottomrule
\end{tabular}%
\caption{
  Prefix tuning (Flat and MLP) and LoRA fine-tuning on T5-Large with different initializations on P3 held-out tasks.
}
\label{tab:table_a01_peft_prefix}
\end{table}

\end{document}